\def\year{2018}\relax
\documentclass[letterpaper]{article} 
\usepackage{aaai18}  
\usepackage{times}  
\usepackage{helvet}  
\usepackage{courier}  
\usepackage{multirow}
\usepackage{booktabs}
\usepackage{amssymb}
\usepackage{amsmath}
\usepackage[table]{xcolor}
\usepackage{amsmath}
\usepackage{float}
\usepackage{verbatim}
\usepackage{url}  
\usepackage{graphicx}  
\begin{document}
%
\title{Investigating Inner Properties of Multimodal Representation and Semantic Compositionality with Brain-based Componential Semantics}
\author{Shaonan Wang$^{1,2}$, Jiajun Zhang$^{1,2}$, Nan Lin$^{3,4}$, Chengqing Zong$^{1,2,5}$ \\
  $^1$ National Laboratory of Pattern Recognition, CASIA, Beijing, China \\ $^2$ University of Chinese Academy of Sciences, Beijing, China \\
  $^3$ CAS Key Laboratory of Behavioural Science, Institute of Psychology, Beijing, China\\
  $^4$ Department of Psychology, University of Chinese Academy of Sciences, Beijing, China \\
  $^5$ CAS Center for Excellence in Brain Science and Intelligence Technology, Shanghai, China  \\
   \{shaonan.wang,jjzhang,cqzong\}@nlpr.ia.ac.cn; linn@psych.ac.cn
}
\setlength\titlebox{2.5in}
\maketitle

\begin{abstract}
Multimodal models have been proven to outperform text-based approaches on learning semantic representations. However, it still remains unclear what properties are encoded in multimodal representations, in what aspects do they outperform the single-modality representations, and what happened in the process of semantic compositionality in different input modalities. Considering that multimodal models are originally motivated by human concept representations, we assume that correlating multimodal representations with brain-based semantics would interpret their inner properties to answer the above questions. To that end, we propose simple interpretation methods based on brain-based componential semantics. First we investigate the inner properties of multimodal representations by correlating them with corresponding brain-based property vectors. Then we map the distributed vector space to the interpretable brain-based componential space to explore the inner properties of semantic compositionality. Ultimately, the present paper sheds light on the fundamental questions of natural language understanding, such as how to represent the meaning of words and how to combine word meanings into larger units.
\end{abstract}

\section{Introduction}
Multimodal models that learn semantic representations using both linguistic and perceptual inputs are originally motivated by human concept learning and the evidence that many concept representations in the brain are grounded in perception \cite{andrews2009integrating}. The perceptual information in such models is derived from images \cite{roller2013multimodal,collell2017imagined}, sounds \cite{kiela2015multi}, or data collected in psychological experiments \cite{johns2012perceptual,hill2014learning,andrews2009integrating}. Multimodal methods have been proven to outperform text-based approaches on a range of tasks, including modeling semantic similarity of two words or sentences and finding the most similar images to a word \cite{bruni2014multimodal,lazaridou2015combining,kurach2017better}.

Despite of their superiority, what happened inside is hard to be interpreted and many questions have been unexplored. For example, it is still unclear 1) what properties are encoded in multimodal representations, and in what aspects do they outperform single-modality representations. 2) Whether different semantic combination rules are encoded in different input modalities, and how different composition models combine inner properties of semantic representations. Accordingly, to facilitate the development of better multimodal models, it is desirable to efficiently compare and investigate the inner properties of different semantic representations and different composition models.

Experiments with brain imaging tools have accumulated evidence indicating that human concept representations are at least partly embodied in perception, action, and other modal neural systems related to individual experiences \cite{binder2011neurobiology}. In summary of the previous work, Binder et al. \shortcite{binder2016toward} propose the ``\textbf{\textit{brain-based componential semantic representations}}'' based entirely on such functional divisions in the human brain, and represent concepts by sets of properties like \textit{vision, somatic, audition, spatial}, and \textit{emotion}. Since multimodal models, in some extent, simulate human concept learning to capture the perceptual information that is nicely encoded in the human brain, we assume that correlating them with brain-based semantics in a proper way would interpret the inner properties of multimodal representations and semantic compositionality.

To that end, we first propose a simple correlation method, which utilizes the brain-based componential semantic vectors \cite{binder2016toward} to investigate the inner properties of multimodal word representations. Our method calculates correlations between the relation matrix given by brain-based property vectors and multimodal word vectors. The resulting correlation score represents the capability of the multimodal word vectors in capturing the brain-based semantic property. Then we employ a mapping method to explore how semantic compositionality works in different input modalities. Specifically, we learn a mapping function from the distributed semantic space to the brain-based componential space. After mapping word and phrase representations to the (interpretable) brain-based semantic space, we compare the transformations of their inner properties in the process of combining word representations into phrases.

Our results show that 1) single modality vectors from different sources encode complementary semantics in the brain, giving multimodal models the potential to better represent concept meanings. 2) The multimodal models improve text-based models on sensory and motor properties, but degrade the representation quality of abstract properties. 3) The different input modalities have similar effects on inner properties of semantic representations when combining words into phrases, indicating that the semantic compositionality is a general process which is irrespective of input modalities. 4) Different composition models combine the inner properties of  constituent word representations in a different way, and the Matrix model best simulate the semantic compositionality in multimodal environment.

\section{Related Work}
\subsection{Investigation of word representations}
There have been some researches on interpreting word representations. Most work investigates the inner properties of semantic representations by correlating them with linguistic features \cite{lingevaluation,yogatamasparse,qiu2016investigating}. Besides, Rubinstein et al. \shortcite{rubinstein2015well} and  Collell and Moens \shortcite{collellimage} evaluate the capabilities of linguistic and visual representations respectively by predicting word features. They utilize the McRae Feature Norms dataset \cite{mcrae2005semantic}, which contains 541 words with a total of 2,526 features such as \textit{an animal, clothing} and \textit{is fast}. These work can be seen as the foreshadowing of our experimental paradigm that correlating dense vectors with a sparse feature space. 

Different from the above work, we utilize the brain-based semantic representations. This dataset contains the basic semantic units directly linked to the human brain, and thus is more complete and more cognitively plausible to represent concept meaning. Furthermore, it is worth noting that all these work does not focus on multimodal representations, and lacks a direct comparison between unimodal representations and multimodal representations. This is exactly our novelty and contribution.

\subsection{Investigation of semantic compositionality} 
Semantic compositionality has been explored by different types of composition models \cite{mitchell2010composition,Dinu_generalestimation,wang2017comparison,wang2017exploiting,wang2017learning,wang2017empirical}. Still, dimensions in many semantic vector space have no clear meaning and thus it is difficult to interpret how different composition models work. Fyshe et al. \shortcite{fyshe2015compositional} tackle this problem by utilizing sparse vector spaces. They use the intruder task to quantify the interpretability of semantic dimensions, which needs manual labeling and the results are not intuitive. Li et al. \shortcite{li2015visualizing} use visualizing methods by projecting words, phrases and sentences into two-dimensional space. This method shows the semantic distance between words, phrases and sentences, but can not explain what happened inside composition. 

The semantic compositionality in computer vision does not receive as much attention as in natural language area. To our best knowledge, the following two studies are most relevant to our work. Nguyen et al. \shortcite{nguyen2014coloring} model compositionality of attributes and objects in the visual modality as done in the case of adjective-noun composition in the linguistic modality. Their results show that the concept topologies and semantic compositionality in the two modalities share similarities. Pezzelle et al. \shortcite{pezzelle2016building} investigate the problem of noun-noun composition in vision. They find that a simple Addition model is effective in achieving visual compositionality. This paper takes a step further, and provides a direct and comprehensive investigation of the composition process in both linguistic and visual modalities. Furthermore, we conduct a pioneer work on multimodal semantic compositional semantics, in which multi-modal word representations are combined to obtain phrase representations. Taken together, our work offers some insights into the behavior of semantic compositionality.

\subsection{Human concept representations and composition}
Classical componential theories of lexical semantics assume that concepts can be represented by sets of primitive features, which are problematic in that these features are themselves complex concepts. Binder et al. \shortcite{binder2016toward} tackle this problem by resorting to brain imaging studies. They propose the ``\textit{\textbf{brain-based componential semantics}}" based entirely on functional divisions in the human brain, and represent concepts by sets of properties like \textit{vision, somatic, audition, spatial}, and \textit{emotion}. The brain-based semantic representations are highly correlated with the brain imaging data, and have been used as an intermediate semantic representations in exploring human semantics \cite{anderson2016predicting}.

There is previous work exploring the question of semantic composition in the human brain \cite{mitchellquantitative,fyshe2015corpora}. To infer how semantic composition works in the brain, they conduct brain imaging experiments of participants viewing words and phrases, and analyze these data by adopting vector-based composition models. Results illustrate that Multiplication model outperforms Addition model on adjective-noun phrase composition, indicating that people use adjectives to modify the meaning of the nouns. Unlike these work, this paper aims to interpret the inner properties of different composition models in achieving compositionality. We hope that the proposed method can feed back into neuroscience to help exploring human concept representations and composition.


\section{Brain-based Componential Semantic Representations}

\begin{figure}[htb]
\includegraphics[scale=0.5]{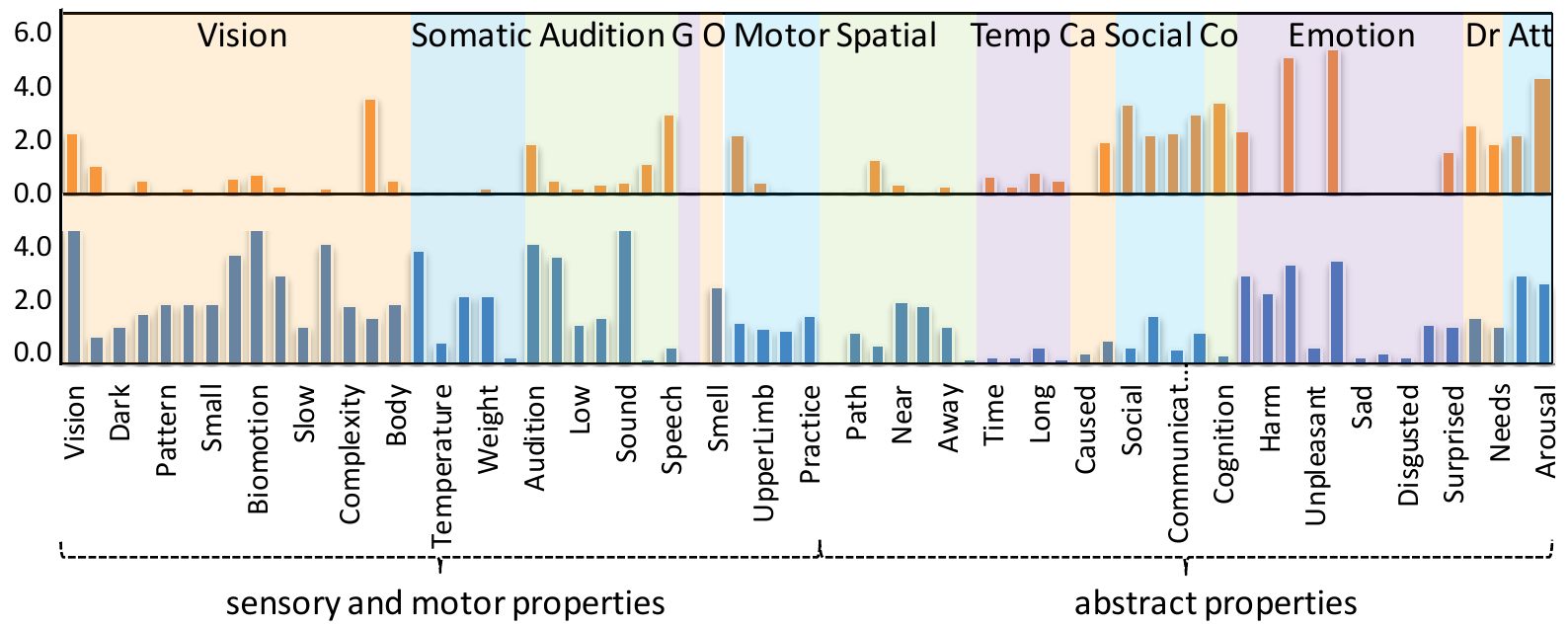}
\centering

\caption{Brain-based componential semantic representations for concepts \textit{happy} (top) and \textit{dog} (bottom). The X-axis denotes attributes (only parts shown) and the Y-axis denotes attribute ratings.}
\end{figure}

The brain-based componential semantic dataset is proposed by Binder et al. \shortcite{binder2016toward}, which contains 535 different types of concepts\footnote{These are 122 abstract words and 413 concrete words including nouns, verbs and adjectives. The dataset can be found at: \url{http://www.neuro.mcw.edu/resources.html}}. Each concept has 14 properties, i.e., \textit{vision, somatic, audition, gustation, olfaction, motor, spatial, temporal, causal, social, cognition, emotion, drive, attention}, and each property contains several attributes (1${\sim}$15). For instance, the \textit{vision} property is described with attributes of \textit{bright, dark, color, pattern, large, small,} etc. Through crowd-sourced rating experiments, each attribute of all 535 concepts is assessed with a saliency score (0${\sim}$6). Figure 1 shows two examples of the brain-based semantic vectors. Consistent with intuition, the concept \textit{happy} as an abstract adjective gets more weights on abstract properties, while the concrete concept \textit{dog} gets more weights on sensory and motor properties. Moreover, via extensive experiments, Binder et al. observe that the brain-based semantic vectors capture semantic similarities and correlate well with the priori conceptual categories, which prove the validity of the dataset.

\section{Inner Properties of Multimodal Representations}
\subsection{Experimental design}
 
\begin{figure}[htb]
\includegraphics[scale=0.45]{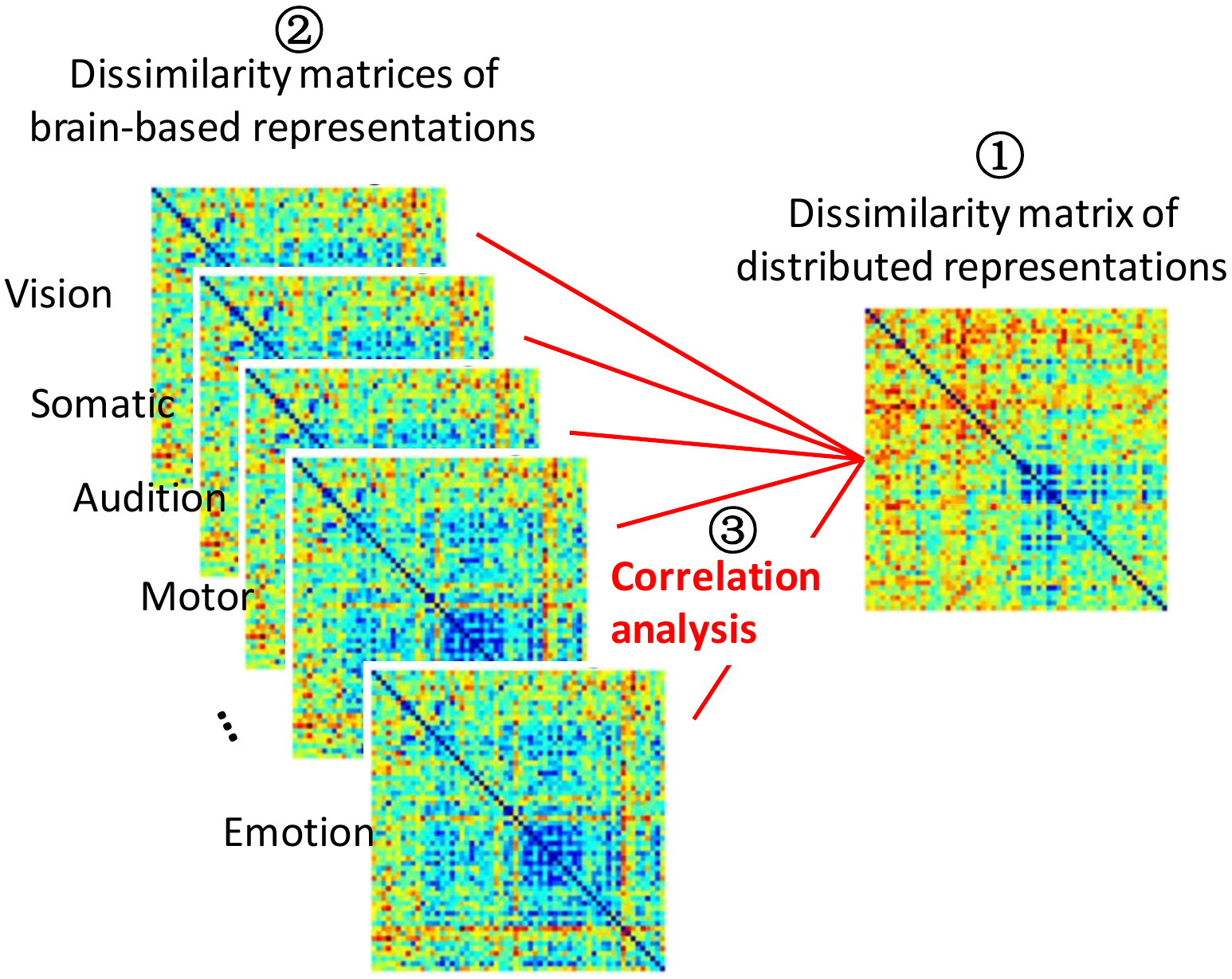}
\centering
\caption{The right dissimilarity matrix is calculated by cosine distance between vectors of each concept pair. The left dissimilarity matrices are calculated by the Euclidean distance between different property vectors of each concept pair. The proposed method calculates the correlations between the dissimilarity matrices given by the brain-based vectors and the distributed representations.}
\end{figure}

To investigate the inner properties of multimodal representations, we adopt the method of representational similarity analysis (RSA) \cite{kriegeskorte2008representational}. As shown in Figure 2, our method involves three steps as follows. (1) For specific distributed representations, we calculate the cosine distance for each word pair in a set of $n$ words (those that appear in both distributed and brain-based vectors), resulting in a dissimilarity matrix with a size of $n\times n$. (2) For brain-based representations, each word corresponds to 14 property vectors. Following Kriegeskorte et al. \shortcite{kriegeskorte2008representational}, we calculate the Euclidean distance\footnote{The metric of cosine similarity can not be adopted here because a few property vectors of certain concepts are zero vectors.} for each word pair (in a set of $n$ words) with different property vectors separately. Each property leads to a dissimilarity matrix and consequently we get 14 dissimilarity matrices. These are $n\times n$ matrices which characterize different semantic aspects of concepts in the brain. (3) We use the Pearson rank correlation coefficient to calculate the relationships between the dissimilarity matrices given by the brain-based vectors and the distributed representations. 


The underlying hypothesis of our method is that if two dissimilarity matrices from different semantic representations have high correlations, then these two representations encode some of the same information. For our method, the two semantic representations are distributed and brain-based property vectors (which characterize the basic semantic aspects of concepts). Therefore, the higher correlation score means that the specific brain-based semantic property is more encoded in the distributed representations.

\subsection{Unimodal and multimodal word representations}

\textbf{Linguistic vectors.} We use the text corpus of Wikipedia 2009 dump\footnote{\url{http://wacky.sslmit.unibo.it}}, which comprises approximately 800M tokens. We discard words that appear less than 100 times and train linguistic vectors by the Skip-gram model \cite{mikolov2013efficient}. We use a window size of 5, set negative number as 5 and iteration number as 3. We finally get 88,501 vectors of 300 dimensions.

\textbf{Visual vectors.} We use visual corpus of ImageNet \cite{deng2009imagenet}, in which we delete words with less than 50 pictures, and sample at most 100 pictures for each word. To extract visual features, we use a pre-trained VGG-19 CNN model\footnote{\url{http://www.vlfeat.org/matconvnet/pretrained/}} and extract the 4096-dimensional activation vector of the last layer. The final visual vectors are averaged feature vectors of multiple images of the same word, which contains 5,523 words of 4096 dimensions.

\textbf{Auditory vectors.} For auditory data, we gather audios from Freesound\footnote{\url{http://www.freesound.org/}}, in which we select words with more than 10 sound files and sample at most 50 sounds for one word. Following Kiela and Clark \shortcite{kiela2015multi}, we use the Mel-scale Frequency Cepstral Coefficient (MFCC) to obtain acoustic features, calculate their “bag of audio words” (BoAW) representations, and obtain the auditory vectors by taking the mean of the BoAW representations of the relevant audio files. We finally get 7,051 vectors of 300 dimensions\footnote{We build auditory vectors with the tool at : \url{https://github.com/douwekiela/mmfeat}}.

\textbf{Multimodal Vectors} To learn multimodal vectors, we choose Ridge   \cite{hill2014multi} and MMskip \cite{bruni2014multimodal}, which are best performing multimodal models. The Ridge model, which utilizes the ridge regression method, first calculates the mapping matrix from linguistic vectors to perceptual vectors, and then predicts the perceptual vectors of the whole vocabulary in linguistic dataset. Finally, the multimodal representations are  concatenation of the $l_2$ normalized predicted vectors and linguistic vectors\footnote{We implement Ridge model with Sklearn (\url{http://scikit-learn.org/}).}, which results in 600-dimensional vectors for 88,501 words. In contrast, the MMskip model injects perceptual information in the process of learning linguistic representations by adding a vision-based objective function\footnote{The MMskip model is implemented with Chainer (\url{http://chainer.org/})}. This objective function is to maximize the distance between positive examples (linguistic vector and its visual vector) and negative examples (linguistic vector and randomly sampled visual vectors). Finally this model gets 88,501 vectors of 300 dimensions.

\subsection{Experimental results}
Based on the proposed correlation method, we first investigate what properties are encoded in different \textit{\textbf{single-modality vectors}}. Next we explore in what properties that \textit{\textbf{multimodal vectors}}\footnote{The multimodal vectors in this paper are calculated with linguistic and visual inputs, because auditory inputs greatly decrease model performance.} perform better than single modality ones, and how they perform on \textit{\textbf{concrete and abstract words}} respectively.

\begin{figure}[htb]
	\includegraphics[scale=0.75]{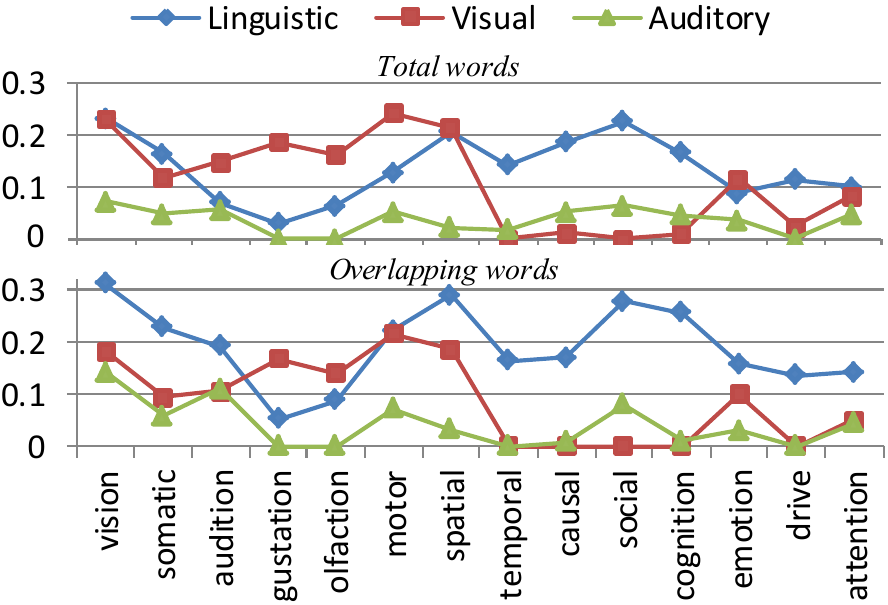}
	\centering
	\caption{Correlations between dissimilarity matrices from single-modality representations and different brain-based property vectors. Top figure shows the results of three single-modality vectors, which cover 530, 202, 436 words of the brain-based vectors respectively. For fair comparison, we show results of overlapping words (188 words -- mostly concrete nouns) in the bottom figure. }
\end{figure}

\subsubsection{Single-modality representations}
Figure 3 shows the inner properties of linguistic, visual and auditory representations, in which the top and bottom figures show the same trends, demonstrating that these vectors encode different semantic aspects of concepts. For instance, linguistic vectors are better at encoding abstract properties like \textit{social} and \textit{cognition}, auditory vectors mainly captures \textit{vision} and \textit{audition} properties, while visual vectors mainly capture properties like \textit{vision}, \textit{motor} and \textit{spatial}. This result indicates that combining different modality inputs has the potential to better represent concept meanings.

\begin{figure}[htb]
	\includegraphics[scale=0.75]{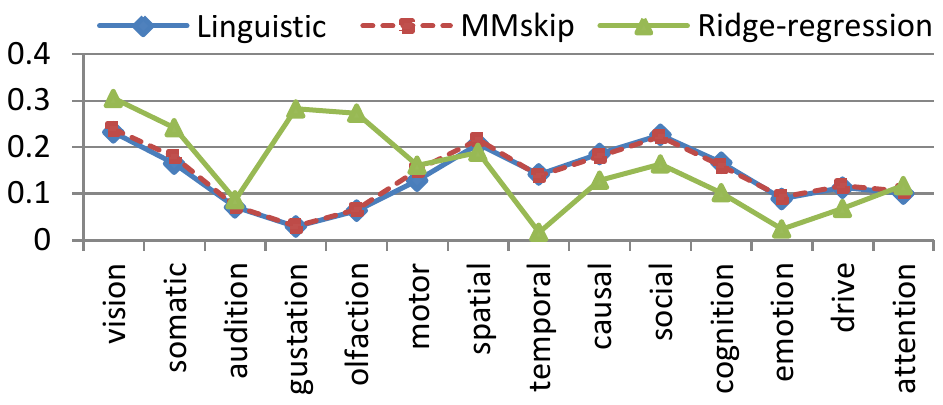}
	\centering
	\caption{Correlation between dissimilarity matrices from the distributed representations and different brain-based property vectors. }
\end{figure}

\subsubsection{Multimodal representations}  
As shown in Figure 4, we can see that compared with linguistic vectors, multimodal vectors from Ridge model have stronger ability on encoding sensory and motor properties but weaker ability on encoding abstract properties. The above results indicate that the visual inputs, which are better at capturing sensory and motor properties, enhance these information conveyed in linguistic representations. On the contrary, the visual inputs contradict abstract properties conveyed in linguistic representations. Especially, the Ridge model achieves the most improvement on \textit{gustation} and \textit{olfaction} properties, because these two properties are significantly captured by the predicted visual vectors. From Figure 4, we can also see that the MMskip model generates multimodal vectors which are similar with (and slightly better than) linguistic vectors. This is because words with visual vectors account for only 5\% of the text corpus.

\begin{figure}[htb]
	\includegraphics[scale=0.75]{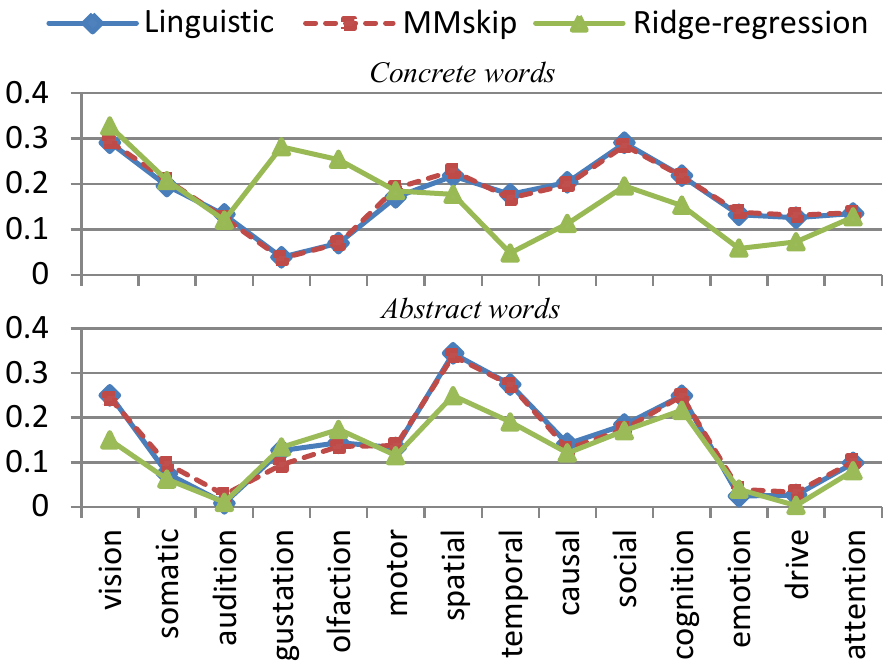}
	\centering
	\caption{Correlations between dissimilarity matrices from distributed representations and different brain-based property vectors on concrete words (top) and abstract words (bottom) respectively.}
\end{figure}

\subsubsection{Concrete vs. abstract words}
 Figure 5 shows the inner properties of semantic representations on concrete and abstract words respectively. It can be seen that both unimodal and multimodal vectors perform differently on concrete and abstract words. For concrete words, they  capture much more inner properties like \textit{vision} and \textit{social}. For abstract words, they encode more inner properties like \textit{spatial} and \textit{cognition}.  Moreover, multimodal vectors achieve lower scores than linguistic vectors on most properties on abstract words. To figure out the reason, we look into the brain-based semantic dataset. We find that abstract concepts have higher attribute scores than concrete concepts on abstract properties (i.e., \textit{spatial, temporal, causal, social, cognition, emotion, drive}, and \textit{attention}), which are poorly captured by visual vectors (the average attribute score is 3.84 and 3.14 respectively). This would lead to performance drop of multimodal models on abstract concepts when mixing with visual inputs. In conclusion, the perceptual input may not be a valuable information for abstract concepts in building multimodal models. 

\section{Inner Properties of Semantic Composition}
\subsection{Experimental design}

\begin{figure}[htb]
	\includegraphics[scale=0.85]{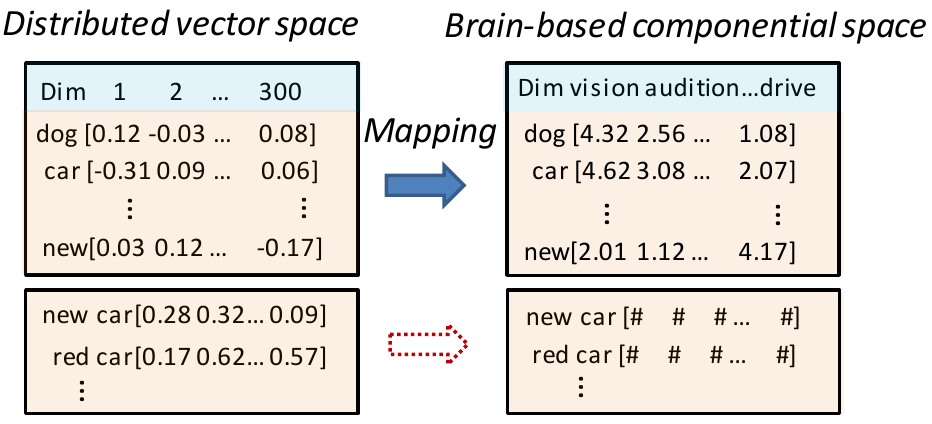}
	\centering
	\caption{Outline of experimental design. The proposed method maps words and phrases in the distributed vector space to the brain-based componential space.}
\end{figure}

To inspect what happened inside semantic compositionality, we design a mapping method to intuitively compare different composition models. The idea behind this method is that via comparing phrase and their constituent word representations in an interpretable vector space, we can observe the changes of inner properties in the process of composition. We hypothesis that there exits a linear/nonlinear map between distributed semantic space and brain-based componential space if the distributed representations implicitly encode sufficient information\footnote{Our experimental results show that linear mapping method works better than non-linear methods. Thus we only report results of linear mapping method.}. 

Figure 6 shows how word and phrase embeddings are mapped to brain-based componential vectors. Specifically, we use $l_2$-normalized word vectors $x$ in distributed vector space and word vectors $y$ in brain-based componential space to learn a mapping function $f: y = f(x)$. Then we map distributed vectors of words and phrases (which are $l_2$ normalized) to the brain-based componential space using the learned mapping function. For linear map, we use the least square method to learn $f$. For nonlinear map, we train a multiple layer perceptron (MLP) neural network. In this paper, we begin our analysis with adjective-noun phrases, where adjectives are used to modify the meaning of nouns. We train the mapping models on the randomly selected 90\% of the words and tune parameters on the left words, in which words include 434 nouns and 39 adjectives in the brain-based semantic dataset.

\subsection{Unimodal and multimodal phrase representations}
\textbf{Visual vectors} This paper chooses the visual genome dataset \cite{krishna2017visual} to learn visual representations, because it contains large annotations of attribute-object pairs (adjective-noun phrases) and their corresponding regions in an image. From this dataset, we extract 2,105,977 adjective-noun pairs. We then delete the phrases which contain adjectives that appear less than 50 times or nouns that appear less than 30 times. To generate phrase vectors, we extract image features with the pre-trained VGG-19 CNN model and calculate the averaged feature vectors of multiple images of the same phrase. Finally we get 4096 dimensional vectors with a vocabulary of 6,874 phrases. 


Based on visual phrase representations, we generate adjective and noun vectors in the same semantic space. Specifically, each word appears in multiple phrases and we calculate the word vectors by averaging their phrase vectors. Finally we get 1,552 word representations.

\textbf{Linguistic vectors} Similar to the linguistic vectors in the previous section, we utilize Skip-gram model and the same text corpus. One difference is that we conduct an extra preprocessing step that combines candidate adjective-noun phrases (i.e., treat phrase as a unit) in the text corpus. This allows the Skip-gram model to generate word and phrase representations simultaneously. For fair comparison, we select the same adjective-phrases as the visual phrases.

\textbf{Multimodal vectors} Since the above linguistic and visual vectors share the same vocabulary, we adopt the concatenation method to generate multimodal word and phrase representations. Specifically, we concatenate the $l_2$ normalized linguistic and visual representations, which results in 600-dimensional vectors for 6,874 phrases and 1,552 words.

\begin{table*}[htbp] \small
	\centering
	\begin{tabular}{ccc}
		\midrule
		\textbf{ Word/Phrase} & \textbf{Visual modality} & \textbf{Linguistic modality} \\
		\midrule
		black & black man, black bag, black top & white, colored,  blue \\
		
		circle & white circles, small circles, red circles & circles, three circles, large circles \\
		
		black circle & circles, round holes, holes & blue circles, red circles, green circles \\
		\midrule
		
		sliver   & steel, shiny, metallic & gold, bronze, gold medal \\
		
		medal   & silver medal, gold medal, red hearts & gold medal, silver medal, bronze \\
		
		silver medal & medal, moon, white circle & gold medal, medal, bronze \\
		
		\midrule
		happy & happy man, funny, young & happy person, happy family, sad \\
		
		face  & white face, round face, clock & faces, white mask, silver mask \\
		
		happy face & facial, facial hair, sad face &  wide eyes, long eyelashes, brown suit \\
		\midrule
	\end{tabular}%
\caption{Top 3 nearest neighbors of an example phrase and its constituent words.}
	\label{tab:addlabel}%
\end{table*}%

\begin{table*}[htbp] \small
	\centering

	\begin{tabular}{llllll} 
		\midrule
		& \textbf{Addition} & \textbf{Multiplication} & \textbf{Matrix} & \textbf{W-addition} & \textbf{Dan} \\
		\midrule
		& Q1\ Q2\ Q3 & Q1\quad Q2\quad Q3 & Q1\  Q2\  Q3 & Q1\  Q2\  Q3 & Q1\  Q2\  Q3 \\
		\midrule
		\textbf{Text}  & 5  36  158 & 1332 3460  5462 & 9  61  227 & 5  36  157 & 15  85  295 \\
		
		\textbf{Image} & 9  28  91 & 1366 3796  5881 & 6  23  66 & 9  28  90 & 8  28  78 \\
		
		\textbf{Multimodal} & 4  33  190 & 1125 3064  5180 & 0  4 26 & 4  32  194 & 2  12  71 \\
		\midrule
	\end{tabular}%
	\label{tab:addlabel}%
		\caption{Rank evaluation of different composition models. The smaller value, the better performance.}
\end{table*}%

\subsection{Composition models}
To investigate how different composition models combine the inner properties of constituent word representations, we make a systematic comparison of five different composition models as follows:

\begin{enumerate} \small 
\item $p_{comp} = Addition(x) = \sum\limits_{i = 1}^n {{x_{i}}} $
\item $p_{comp} = Multiplication(x) = \prod\limits_{i = 1}^n {{x_{i}}} $ 
\item $p_{comp} = W\text{-}addition(x) = \sum\limits_{i = 1}^n {f({W_v}{x_{i}})} $
\item $p_{comp} = Matrix(x) = \sum\limits_{i = 1}^n {f({W_m}{x_{i}})} $
\item $p_{comp} = Dan(x) = f({W_d}(\sum\limits_{i = 1}^n {{x_{i}}} ) ), $
\end{enumerate}
where $x_i$ denotes word representations, $n=2$ is the number of words in a phrase, and \small $\{W_v, W_m, W_d\} $ \normalsize $\in \mathbb{R}^{d \times d}$ are trainable parameters. The nonlinear activation function $f$ used here is  $tanh$. 

Following Diam \shortcite{dima2015reverse}, we adopt a mean square error (MSE) objective function to estimate the modal parameters:

\begin{equation}
 J = min({\left\| {{p_{comp}} - {p_{gold}}} \right\|^2}  +  {\lambda _1}({\left\| {{W_x}} \right\|^2}),
\end{equation}
where $p_{comp}$ is the compositional phrase vector calculated by composition models, and $p_{gold}$ is the gold phrase vector that directly learned from data. Moreover, we use regularization coefficient $\lambda _1$ on model parameters $\{W_v, W_m, W_d\} $. In the experiment, the phrase vectors are randomly partitioned into training, testing and development splits in 7:2:1. Note that we do not train the embedding vectors along with the composition models. Although this could potentially benefit the results, we aim to explore the effects of different composition models in different input modalities.

\subsection{Experimental results}
To intuitively show the characteristic of learned \textit{\textbf{word and phrase representations}} in visual and linguistic modalities, we calculate their nearest neighbors using cosine similarity. Based on the proposed mapping method, we first investigate the inner properties of \textit{\textbf{semantic compositionality}} in linguistic and visual modalities respectively. Next we employ a quantitative analysis to inspect the ability of different \textit{\textbf{composition models}} in capturing the composition rules contained in different modality inputs. After that we explore the effects of different composition models on  \textit{\textbf{multimodal compositional semantics}}. Finally, we show \textit{\textbf{an example}} to see the inner properties changes in combining words into phrases.

\subsubsection{Word and phrase representations} 
As shown in Table 1, the semantic representations in linguistic and visual modalities show different characteristics. In visual modality, words and phrases with similar shape are nearest neighbors, such as \textit{black circle} and \textit{holes}, \textit{face} and \textit{clock}. Moreover, the nearest neighbors of a word in visual modality are sometimes the phrases that begin with this word,  for example the nearest neighbors of \textit{black} are \textit{black man, black bag, black top}. This is because visual word vectors are calculated as the averaged phrase vectors. As in linguistic modality, semantic representations are learned from text corpus, thus there are morphological similar words group together like \textit{circle} and \textit{circles}, \textit{face} and \textit{faces}. There are also nearest neighbors which are semantic related phrases, such as \textit{happy face} with its nearest neighbors of \textit{wide eyes} and \textit{long eyelashes}.

\begin{figure*}[htb]
	\includegraphics[scale=0.85]{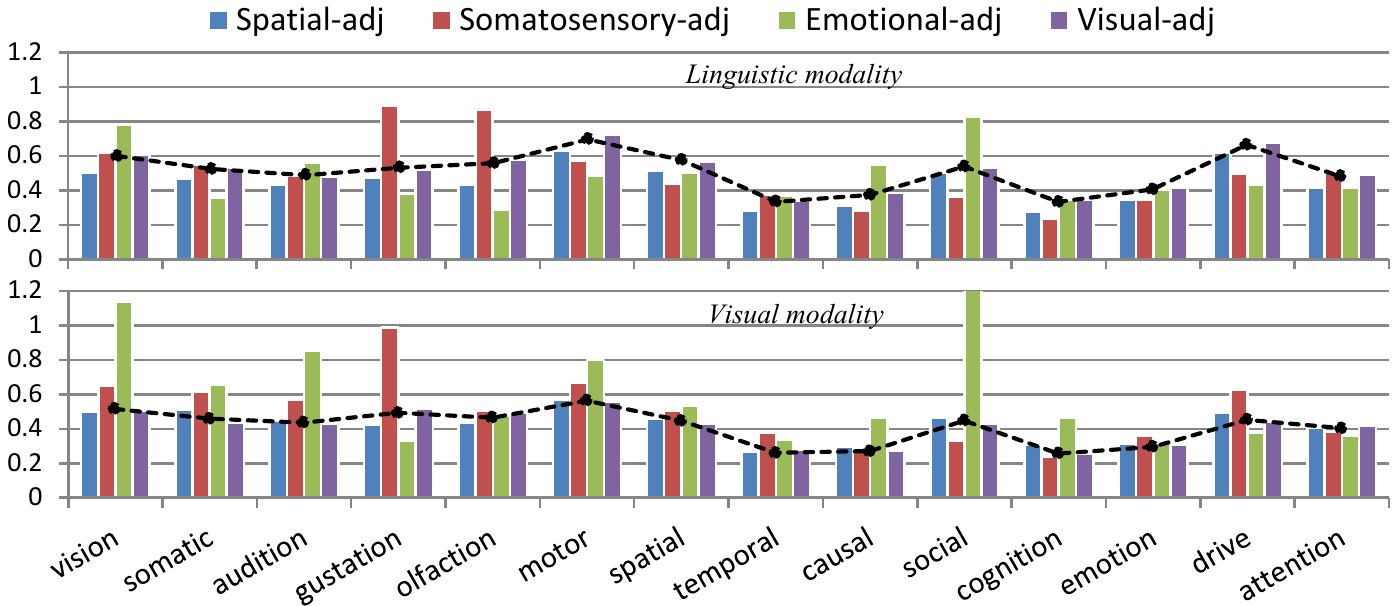}
	\centering
	\caption{Mean property difference between nouns and its adjective-noun phrases (in 4 categories) in linguistic modality (top) and visual modality (bottom). The black dotted line show the average value on all category phrases.}
\end{figure*}

\subsubsection{Semantic compositionality}
To investigate the inner properties of semantic compositionality contained in linguistic and visual inputs, we adopt the proposed mapping method to compare the representations of nouns and its adjective-noun phrases in brain-based componential space. For fine-grained analysis, we divide the adjectives into four categories: spatial (e.g., \textit{small, big}), somatosensory (e.g., \textit{hot, heavy}), visual (e.g., \textit{white, shiny}), and emotional (e.g., \textit{happy, angry}). 

Figure 7 shows the absolute mean property difference\footnote{In this paper, each property contains several attributes and the property difference is its average attribute difference.} between nouns and its adjective-noun phrases. We can see that linguistic and visual modalities show the same characteristic: the adjectives mostly affect properties of \textit{vision, motor, social}, and \textit{drive}. This indicates that semantic compositionality is a general process which is irrespective of input modalities. Another observation is that different adjectives have different effects on semantic compositionality. For example, the emotional adjectives have the greatest impact on the inner properties of their modified nouns, especially on \textit{social, vision, audition} and \textit{motor} properties. The somatosensory adjectives mostly influence  \textit{gustation, olfaction, vision} and \textit{somatic} properties, while the visual and spatial adjectives mostly influence \textit{motor} and \textit{drive} properties.

\subsubsection{Composition models}
To compare different composition models in unimodal and multimodal environment, we employ the rank evaluation method \cite{dima2015reverse} which calculates the rank of similarity between a predicted phrase vector and its gold phrase vector in similarity between the predicted phrase vector and vectors of all phrase vocabulary. Specifically, we compute the first, second and third quartiles (Q1, Q2, Q3) across the test phrases. A Q1 value of 2 means that the first 25\% of the data is only assigned ranks 1 and 2 (i.e., the phrase vectors predicted by the first 25\% of data are all most or second most similar to their corresponding gold phrase vectors). Similarly, Q2 and Q3 refer to the ranks assigned to the first 50\% and 75\% of data, respectively. 

As shown in Table 2, the Addition model achieves the best result on linguistic modality, and the Matrix model obtains the best performance on visual and multimodal modalities. The Multiplication model, which is considered to be the most appropriate strategy for human semantic compositionality \cite{mitchellquantitative}, is not suitable for our distributed representations. Furthermore, we can see that composition models perform better in multimodal environment, indicating that multimodal information provides a better ground for semantic compositionality.

\subsubsection{Multimodal compositional semantics}
Based on the proposed mapping method, we calculate the attribute difference between representations of nouns and its adjective-noun phrases in brain-based componential space. We find that different composition models have different effects. Take the composition of \textit{old man} (in multimodal environment) for example, the Addition model gets lower values on attributes of \textit{biomotion, body, speech, etc.} and higher values on \textit{temporal} related attributes, while the Multiplication model achieves lower values on attributes like \textit{biomotion, face} and \textit{body}, and higher values on attributes like \textit{colour, scene} and \textit{time}. 

To further investigate the effects of different composition models on multimodal compositional semantics, we divide the nouns into 7 categories: place (e.g., \textit{street, mountain}), human (e.g., \textit{boy, family}), animal (e.g., \textit{bird, dog}), body part (e.g., \textit{hair, eye}), tool (e.g., \textit{glass, football}), vehicle (e.g., \textit{car, truck}), and food (e.g., \textit{cheese, coffee}). Together with the four kinds of adjectives, we divide all phrases in brain-based dataset into 19 categories\footnote{We use the category annotations in brain-based semantic dataset (Binder et al. 2016). Specifically, we select adjective categories that contain more than 5 words and noun categories that contain more than 10 words.}. For each category of phrases, we compute its absolute mean difference between nouns and its adjective-noun phrases on all brain-based semantic attributes, in which phrase representations are combined by different composition models.

\begin{figure}[htb]
	\includegraphics[scale=0.8]{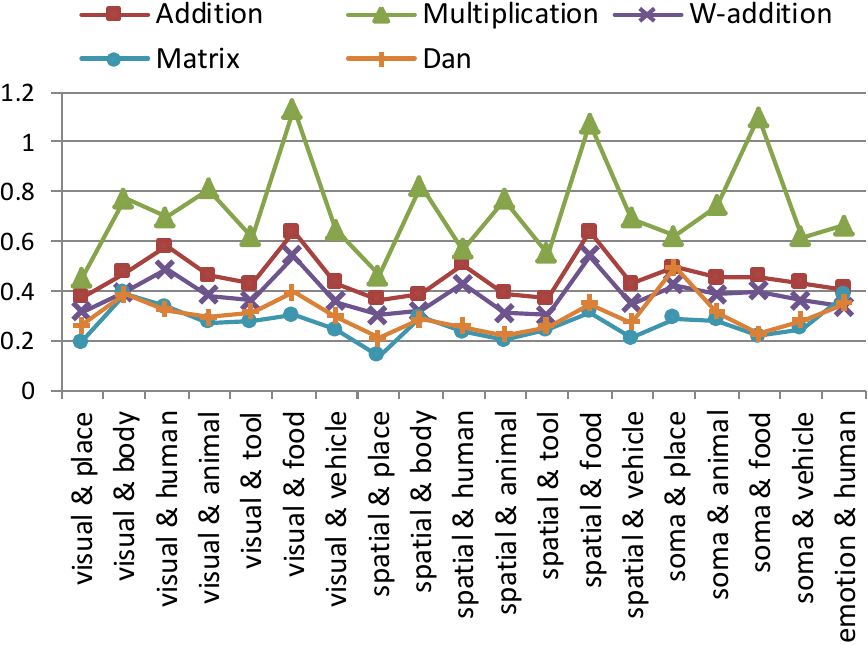}
	\centering
	\caption{Mean property difference between nouns and its adjective-noun phrases (in 19 categories) in multimodal environment, in which phrase representations are obtained by 5 different composition models.}
\end{figure}

\begin{figure*}[htb]
	\includegraphics[scale=0.9]{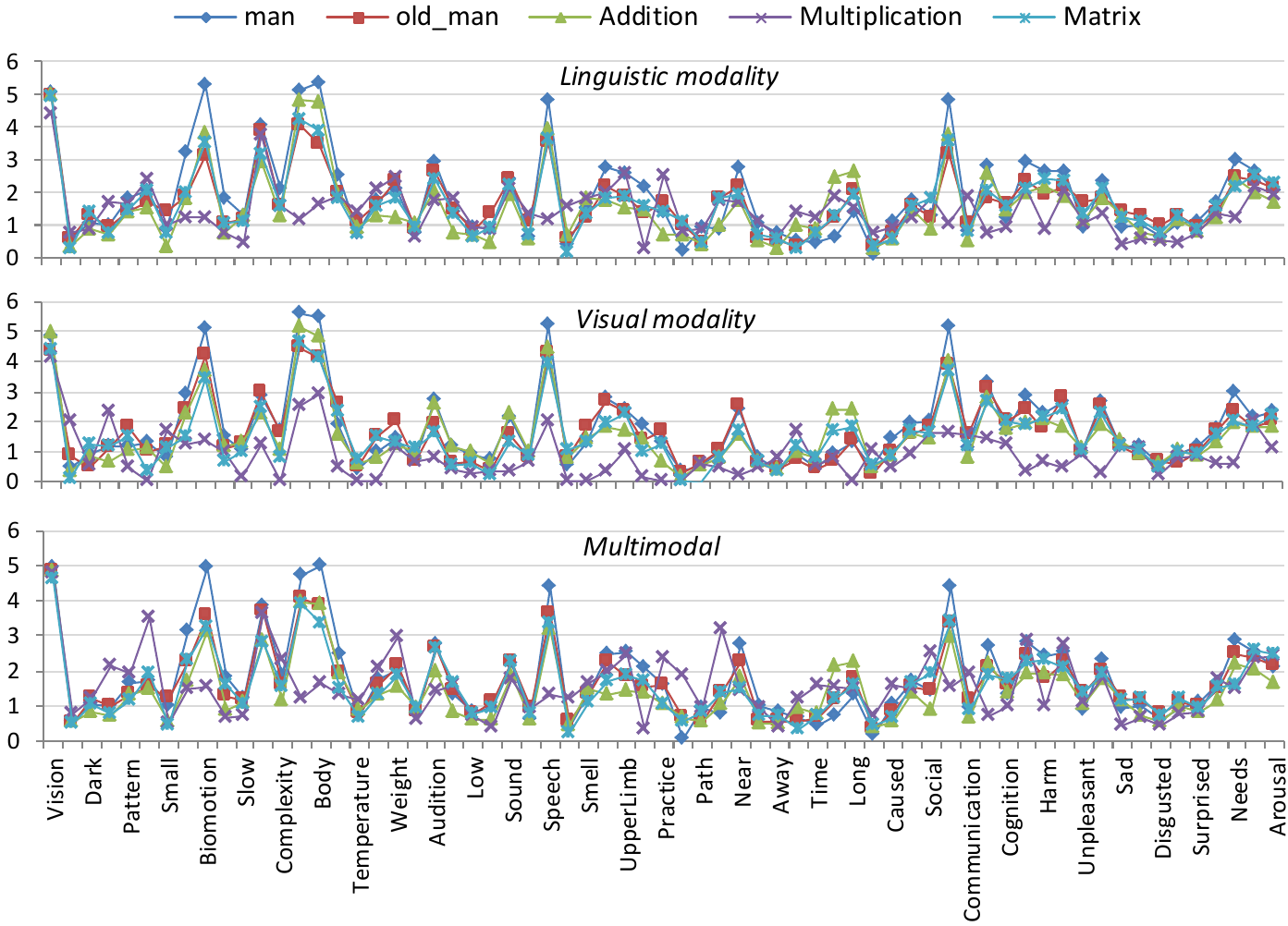}
	\centering
	\caption{Attribute ratings for \textit{man}, \textit{old man} directly extracted from data and \textit{old man} calculated by composition models in linguistic (top), visual (midddle) and multimodal (bottom) environment.}
\end{figure*}

As shown in Figure 8, the composition models with parameters (i.e., Matrix, W-addition, Dan) achieve smaller values than the models without parameters (i.e., Addition, Multiplication), in which Matrix model achieves the smallest value. In other words, the phrase vectors predicted by the Matrix model are most similar with their constituent noun vectors. This result indicates that the composition models with parameters put more importance weights on nouns in composition of adjective-nouns phrases.

\subsubsection{An example}
Figure 9 shows an example word \textit{man} and phrase \textit{old man} in brain-based componential space, which are mapped from distributed vector space with the proposed mapping method. The ``old\_man" line in the figure, which is the representation of phrase \textit{old man} directly extracted from the corpus, can be seen as the standard phrase representations, and they show the similar trend in linguistic, visual and multimodal environment. Nevertheless, there are slight differences. For instance, in linguistic modality, the \textit{old man} achieve higher values on \textit{long, duration, time, landmark}, etc. attributes, while in visual modality the \textit{old man} achieve higher values on \textit{pattern, weight, texture}, etc. attributes. 

The Dan model and W-addition model have similar characteristics with Matrix and Addition model respectively, which we do not shown in the figure for clarity. The three different composition models in Figure 9 shows different characteristics. The Addition model gets higher value on attributes like \textit{duration, long, time, number, sad, taste}, and lower vlaue on attributes like \textit{biomotion, motion, human, head, upperlimb, speech}. The Multiplication model obtains higher value on attributes like \textit{bright, color, small, number, time, communication}, and lower value on attributes like \textit{biomotion, face, human, body, speech}. The Matrix model gets higher value on attributes like \textit{scene, duration, social, long, pain, cognition}, and lower value on attributes like \textit{biomotion, body, human, speech, face}. Taken together, we conclude that different composition models have different effects on inner properties of semantic representations. 

\section{Conclusion and Future Work}
In this paper, we utilize the brain-based componential semantics to investigate what properties are encoded in semantic representations and how different composition models combine meanings. Our results shed light on the potential of combing representations from different modalities, building better multimodal models by distinguishing different types of concepts, and learning semantic compositionality in multimodal environment.

\section*{Acknowledgements}
The research work is supported by the National Key Research and Development Program of China under Grant No. 2017YFB1002103, the Natural Science Foundation of China under Grant No. 61333018, and the Strategic Priority Research Program of the CAS (Grant XDB02070007).

\fontsize{9.5pt}{10.5pt} \selectfont
\bibliography{aaai}
\bibliographystyle{aaai}

\end{document}